%% file: paper.tex
\documentclass[conference]{IEEEtran}
\IEEEoverridecommandlockouts
\usepackage{amsmath,amssymb,amsfonts}
\usepackage{algorithmic}
\usepackage{graphicx}
\usepackage{textcomp}
\usepackage[numbers, sort]{natbib}
\usepackage[x11names, dvipsnames]{xcolor}
\usepackage[hidelinks]{hyperref}
\usepackage{listings}
\usepackage[T1]{fontenc}
\usepackage{booktabs}
\usepackage{slantsc}
\usepackage{multirow}
\usepackage{multicol}
\usepackage{soul}
\usepackage{bbm}
\usepackage{stfloats}
\usepackage{flushend}

\newcommand{\dfmit}{DF\textsubscript{MIT}}
\newcommand{\dffrac}{DF\textsubscript{Frac}}
\newcommand{\rankdel}{Rank\textsubscript{Del}}
\newcommand{\selfexp}{\textsc{SelfExp}}
\newcommand{\selfexpk}{\textsc{TopK}}
\newcommand{\topk}{top-$k$}
\let\vec\mathbf

\begin{document}

\title{Can Large Language Models Explain Themselves? A Study of LLM-Generated Self-Explanations
\thanks{\hspace{-1.38em} $^*$\,Equal contribution.}
}

\author{
\begin{tabular}{ccccc}
    Shiyuan Huang$^*$ & Siddarth Mamidanna$^*$ & Shreedhar Jangam & Yilun Zhou & Leilani H. Gilpin \\
    UC Santa Cruz & UC Santa Cruz & UC Santa Cruz & MIT CSAIL & UC Santa Cruz\\
    \multicolumn{5}{c}{\{shuan101, spmamida, sjangam\}@ucsc.edu, yilun@csail.mit.edu, lgilpin@ucsc.edu}
\end{tabular}
}

\maketitle

\begin{abstract}
Large language models (LLMs) such as ChatGPT have demonstrated superior performance on a variety of natural language processing (NLP) tasks including sentiment analysis, mathematical reasoning and summarization. Furthermore, since these models are instruction-tuned on human conversations to produce ``helpful'' responses, they can and often will produce explanations along with the response, which we call \textit{self-explanations}. For example, when analyzing the sentiment of a movie review, the model may output not only the positivity of the sentiment, but also an explanation (e.g., by listing the sentiment-laden words such as ``fantastic'' and ``memorable'' in the review). How good are these automatically generated self-explanations? In this paper, we investigate this question on the task of sentiment analysis and for feature attribution explanation, one of the most commonly studied settings in the interpretability literature (for pre-ChatGPT models). Specifically, we study different ways to elicit the self-explanations, evaluate their faithfulness on a set of evaluation metrics, and compare them to traditional explanation methods such as occlusion or LIME saliency maps. Through an extensive set of experiments, we find that ChatGPT's self-explanations perform on par with traditional ones, but are quite different from them according to various agreement metrics, meanwhile being much cheaper to produce (as they are generated along with the prediction). In addition, we identified several interesting characteristics of them, which prompt us to rethink many current model interpretability practices in the era of ChatGPT(-like) LLMs.
\end{abstract}

\begin{IEEEkeywords}
Machine Learning, Natural Language Processing, Interpretability, Large Language Model (LLM)
\end{IEEEkeywords}

\section{Introduction}
Over the past couple years, instruction-tuned \citep{ouyang2022training} large language models (LLMs) have revolutionized the field of natural language processing (NLP) with its superior performance on a wide variety of tasks and benchmarks such as sentiment analysis \citep{wang2023chatgpt}, mathematical reasoning \citep{frieder2023mathematical} and summarization \citep{yang2023exploring}. Commercial models such as ChatGPT and GPT-4 by OpenAI, Bard by Google and Claude by Anthropic have been used to (partially) automate many human labor-intensive processes such as education \citep{kohnke2023chatgpt}, trip planning \citep{wong2023autonomous} and customer service \citep{george2023review}. A notable feature of these models is the ability to provide (seemingly) reasonable explanations. For example, when asked to solve a math problem, they often include detailed derivation steps even when not explicitly instructed to do so. Similarly, when asked to analyze the sentiment of a movie or book review, they often spontaneously explain their decision with supporting evidence, such as by highlighting sentiment-laden words or explaining the sarcasm in the text.

How should we think about these explanations? In particular, do these explanations faithfully support the model's prediction (e.g., the final answer of the math problem or the positive/negative judgment in sentiment analysis), or do they simply give a sense of transparency while being decoupled from the prediction? To what extent are they similar to or different from the explanations generated from traditional techniques \citep[e.g.][]{simonyan2013deep, li2015visualizing, zeiler2014visualizing, li2016understanding, ribeiro2016should, lundberg2017unified}, which are commonly used to study predecessor models such as BERT \cite{devlin2019bert} and GPT-2 \cite{radford2019language}? For any ``negative results'' about the effectiveness of these explanations, should we attribute them to the model's inability to generate the explanations, or the limitations in our evaluations \citep{deyoung2020eraser, disagreementProblem} about them?

\begin{figure}[t]
    \centering
    \includegraphics[width=\columnwidth]{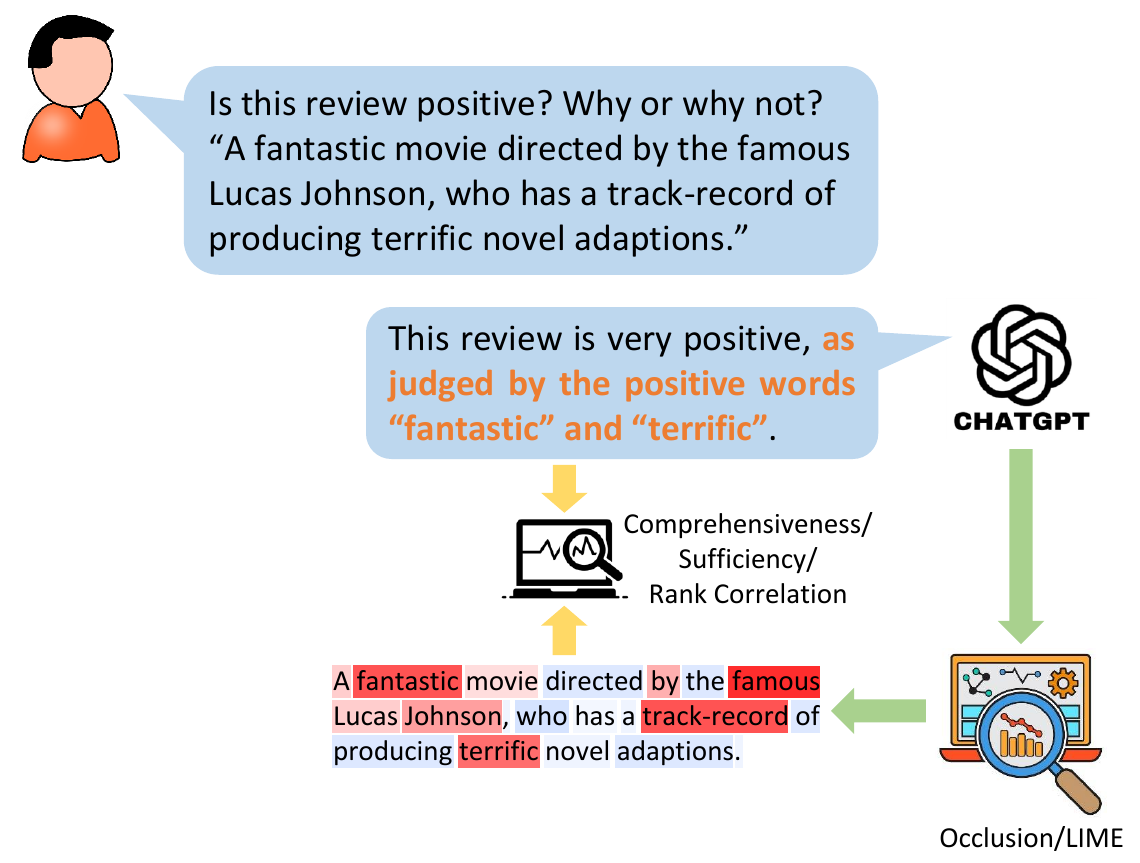}
    \caption{An overview of our investigation. Current conversational LLMs can explain their answers (e.g., by highlighting important words in the input), often automatically or at least when asked to. How should we think of these \textit{self-explanations}? In this paper, we study them in relationship to traditional model interpretability techniques such as occlusion saliency and LIME, and on various metrics such as comprehensiveness, sufficiency and rank agreement. Our findings suggest that we may need to rethink the model interpretability pipeline for analyzing these models.}
    \label{fig:1}
\end{figure}

In this paper, we seek answers to the above questions, using OpenAI's ChatGPT as the model, feature attributions as the explanation type and sentiment analysis as the task. All three design choices are made based on popularity in the respective aspect: ChatGPT is arguably the most widely used instruction-tuned LLM, feature attribution (also known as saliency map or feature importance) explanation is by far the most popular type of explanations studied in the interpretability literature, and sentiment analysis is often the first task to consider when benchmarking these explanations due to the intuitive alignment between strong sentiment words (e.g, ``fantastic'' or ``terrible'') with feature attribution scores. 

In the experiment, we study two paradigms of generating the explanations: generating the explanation first before the prediction (explain-then-predict or E-P), versus generating the prediction first and then explaining it (predict-and-explain or P-E)  \cite{eSNLINLEs}. In each paradigm, we construct two methods to eliciting the explanations: asking the model to generate a full list of feature attribution explanations containing the importance score of every word, and asking the model to just return the top-$k$ most important words. We then compare them with two classic interpretability methods, occlusion \cite{li2016understanding} and LIME \cite{ribeiro2016should}. The comparison is done on two fronts, a suite of faithfulness evaluation metrics \cite{zhou2022solvability} and a set of disagreement measurements among explanations \cite{disagreementProblem}. 

Quantitatively, the model's self-explanations perform on par with traditional methods via faithfulness evaluations, but at the same time differ greatly from them (on various agreement metrics). More importantly, we find that the classic interpretability pipeline of defining and evaluating model explanations may be fundamentally ill-suited for these LLMs with quite human-like reasoning ability. Specifically, the model does not produce sufficiently fine-grained model predictions for us to meaningfully compute and evaluate explanations, in ways that we have been done for predecessor models such as BERT \citep{devlin2019bert}. Nonetheless, we believe that this should be considered as a limitation on the current interpretability practice, rather than the model, as many such computations (e.g., identifying the effect of removing a single word in a long sentence) are ill-posed for humans as well. Overall, we hope that this work could bring formal investigations of LLM-generated self-explanations to the attention of the community, and inspire more work towards better understanding and using them.

\section{Related Work}
Our research lies at the intersection of three areas within interpretability research: feature attribution explanations, evaluations of feature attribution explanations and LLM-generated self-explanations. The former two areas are well-studied, so we give emphasis to works that are directly used in our paper. For broader contexts and more comprehensive coverage, we refer the reader to the textbook by \citet{molnar2022}. The last area is relatively new, and while, to the best of our knowledge, we are the first to rigorously study LLM-generated feature attribution explanations, we briefly discuss some related ideas. 

\subsection{Feature Attribution Explanations}
\label{sec:fa-methods}
At a high level, feature attribution explanations assign attribution scores to input features \citep{zhou2021feature}. There are in general two ways of defining input features. The first is based on the model's response to infinitesimal perturbation of the input feature value, with representative methods such as the gradient saliency \citep{simonyan2013deep}, smooth gradient \citep{smilkov2017smoothgrad} and integrated gradient \citep{sundararajan2017axiomatic}. For NLP models, the perturbation is often done in the embedding space \citep{li2015visualizing}. However, since ChatGPT is a closed-source model that does not provide input gradients, none of these explanations are possible to calculate. 

Instead, we focus on the second approach: defining feature importance based on a model's response to inputs with various features removed. The simplest definition is the occlusion saliency \citep{zeiler2014visualizing, li2016understanding}, with the importance of a feature (i.e., word) defined as the change in the model prediction on the sentence with that word removed. To capture non-linear interactions among different words, LIME \citep{ribeiro2016should} and SHAP \citep{lundberg2017unified} remove multiple features at the same time and define feature importance as linear regression coefficients or the game-theoretic Shapley value \citep{roth1988shapley} respectively. The word removal strategy can be influential, with \citet{pham2021double} finding that some strategies favor some evaluations. In this paper, we adopt the mainstream approach of word deletion (i.e., removing the word ``This'' from the sentence ``\textit{This is a great movie}'' results in ``\textit{is a great movie}''), and leave the exploration of other strategies (e.g., mask infilling) to future work. 

\subsection{Evaluations of Feature Attribution Explanations}

Our main contribution of this paper is an investigation into the relative strengths and weaknesses of LLM-generated self-explanations compared to traditional explanations such as occlusion and LIME. In this paper, we evaluate both their faithfulness and agreement with each other.

For faithfulness evaluations, we employ several popular metrics, including comprehensiveness \citep{deyoung2020eraser}, sufficiency \citep{deyoung2020eraser}, decision flip rate under the removal of the most important token/word \citep{chrysostomou2021improving}, fraction of tokens need to be removed for decision flip \citep{serrano2019attention} and deletion rank correlation \citep{alvarez2018robustness}. They all subscribe to a fundamental principle -- modifying a significant feature, as identified by the explanation, should markedly influence the prediction -- but implement it differently.

\begin{table*}[b]
\input{tables/ep}

Furthermore, recognizing that there is no ``ground truth'' model reasoning available (due to the black-box nature of these models), one type of evaluation modifies the dataset and retrains the model~\citep{adebayo2020debugging, zhou2021feature, adebayo2021post, yang2019benchmarking, bastings2021will}, such that the model has to use a specific type of reasoning for high accuracy, and then evaluates whether the saliency maps of such high-accuracy models align with the expectation. Given the high cost and limited flexibility (API access only) of finetuning, we leave these evaluations to future work.

We also study the correlation between the LLM self-explanation and traditional techniques using the set of disagreement metrics introduced by \citet{disagreementProblem}. Additional manifestation of the disagreement problem includes the finding that the explanations themselves are often unstable \citep{ghorbani2019interpretation} and that they are very sensitive to hyperparameter choices \citep{bansal2020sam}. 

\subsection{LLM-Generated Self-Explanations}

Generative models, those that can generate texts such as the GPT family of models \citep{radford2018improving, radford2019language}, offer a distinct way of explaining them -- by simply generating the explanations directly. \citet{eSNLINLEs} trained models on supervised pairs of labels and explanations and considered two orderings of prediction and explanation: making a prediction and then explaining it, and generating an explanation and using it to make a prediction. We study both approaches in our paper. In addition, \citet{eSNLINLEs} found that training the model to generate the explanation, both before and after the prediction, results in a slight accuracy drop, casting doubt on the inherent usefulness of the explanation to the prediction.

For ChatGPT and later models, chain-of-thought generation \citep{chainOfThoughtWReasoning}, which can be considered as the model's explanation, is often helpful for the accuracy of the final answer, especially on complex reasoning tasks such as solving math problems \citep{hendrycks2021measuring}. This evidence implies that the model could meaningfully reason from such explanations. By contrast, \citep{chen2023models} studied properties of self-generated counterfactual explanations, proposed several metrics and found that these explanations are often misleading and lead to the wrong mental model of human inspectors.

\section{Methodology}

Our main contribution in this paper is a suite of systematic analyses of LLM-generated self-explanations in the sentiment analysis domain. We first covers the basics of auto-regressive LLMs (such as ChatGPT and similar conversational models), and then describe our prompting strategy and considerations. After that, we introduce the two traditional interpretability methods that we compare to, and also evaluation metrics.

\subsection{Auto-regressive LLMs}
Auto-regressive LLMs, as pioneered by GPT-1 \citep{radford2018improving} and GPT-2 \citep{radford2019language}, operate by successively generating the next word (or token) given the history (called context window), until a special stop token is generated. The generation can be stochastic or deterministic, and various decoding strategies could be used to improve the quality of the generation \citep{holtzman2019curious}. In all of our experiments, we use deterministic greedy decoding (i.e., with temperature $=0$). Thus, the generation process can be modeled as the function 
\begin{align}
    \vec x_g = g(\vec x_c), 
\end{align}
where $\vec x_c$ is the sequence of context words, and $\vec x_g$ is the sequence of generated words. 

A conversational LLM (i.e., an LLM that can engage in human conversations) typically has three message types: an initial ``system'' message to set the general tone and instruction of the conversation, followed by rounds of interleaving ``user'' and ``assistant'' messages. While the exact details of ChatGPT is unknown due to its proprietary nature, similar open-source models such as Alpaca \citep{taori2023alpaca} and Vicuna \citep{vicuna2023} all adopt the same practice to represent the message history: by concatenating individual message prepended with the role indicator. 

For example, if the model (i.e., assistant) needs to answer the user's second question below: 

\vspace{0.08in}
\noindent
\fbox{%
    \parbox{0.96\columnwidth}{
    \textit{System: You are a math teacher.\\
User: What is one solution to x\^{}2=1?\\
Assistant: x=1 is a solution.\\
User: Are there other solutions?}
    }
}
\vspace{0.08in}

\noindent the context $\vec x_c$ is constructed to be the following text: 

\vspace{0.08in}
\noindent
\fbox{%
    \parbox{0.96\columnwidth}{
    \textit{System: You are a math teacher. User: What is one solution to x\^{}2=1? Assistant: x=1 is a solution. User: Are there other solutions? Assistant:}
    }
}

\vspace{0.08in}

\noindent which is a concatenation of all of the past messages.

\begin{table*}[t]
\input{tables/pe}

\begin{table*}[b]
\input{tables/wild}

\begin{table*}[t]
\input{tables/epk}

\begin{table*}[b]
\input{tables/pek}

\subsection{Prompting}
\label{sec:method-prompting}

Our central goal is to have the model generate a prediction on the sentiment of the input sentence and a feature attribution explanation for the prediction. We include both explanation pipeline setups studied by \citet{eSNLINLEs}, \textit{explain-then-predict} (E-P), in which the model first generates the explanation and then arrive at a prediction based on it, and \textit{predict-and-explain} (P-E), in which the model first generates the prediction and then explains it.

Our prompts for E-P and P-E setups are shown in Table \ref{tab:ep-prompt} and \ref{tab:pe-prompt} respectively. In the design of both prompts, we have several considerations. First, in order to minimize the number of turns in the conversation, we instruct the model to output both the explanation and prediction at the same time, but in clearly defined ordering. Second, in some of our experiments, we need to feed the model with partial inputs (i.e., sentences with some words removed) such as ``\textit{This \st{is} a really \st{great} movie with \st{beautiful} music and intricate \st{plot}}'' $\Rightarrow$ ``\textit{This a really movie with music and intricate}''. We empirically find that models without any guidance sometimes will refuse to provide an answer, so we intentionally ask them to give their best guess despite input ambiguity. Finally, we initially had significant difficulty making the model to output the word attribution values in a consistent format, but in the end find that the ``Python'' format works the best for the model to follow. 

Despite the success in generating syntactically correct outputs, a concern with this Python format guiding is that the unnaturalness of output (compared to, e.g., Table \ref{tab:top-k-in-the-wild}) may result in the explanation values also being less faithful to the model's true reasoning. Thus, to remove any effect of the unnatural prompt, and recognizing that humans, to which these models are aligned \citep{ouyang2022training}, are not be able to produce detailed attribution scores, we additionally consider two \topk{} versions of the E-P and P-E setups.  We construct these prompts to be as natural as possible while ensuring that the result remains easy to parse, and the final prompts are shown in Table \ref{tab:ep-prompt-topk} and \ref{tab:pe-prompt-topk} respectively. The number $k$ is chosen dynamically according to each input sentence as $\min(1, \lfloor L / 5\rfloor)$, where $L$ is the number of words in the sentence.

For each prompt, the model generates a decimal point number for the prediction confidence. Since we do not have model weight access, we cannot compute the logit or probability of any generated token. Thus, we take this decimal point (in the range of 0.0 to 1.0) as the model's prediction, which is used to evaluate the model's accuracy. 

\subsection{Traditional Interpretability Methods}

We benchmark the LLM-generated self-explanations with traditional interpretability methods that are often used to explain classifiers. As mentioned in Section~\ref{sec:fa-methods}, gradient-based attribution values cannot be used due to the lack of input gradient access. Instead, we consider two perturbation-based methods, occlusion \citep{li2016understanding} and LIME \citep{ribeiro2016should}. The occlusion method defines the attribution value on word $w$ as 
\begin{align}
    a(w)=f(\vec x)-f(\vec x_{-w}), 
\end{align}
where $f(\cdot)$ is the model prediction (as the generated decimal point number in the range of 0.0 to 1.0) for the input sentence, $\vec x$ is the original sentence, and $\vec x_{-w}$ is the input sentence with the word $w$ removed. Intuitively, the occlusion value captures the effect of a single word removal on model prediction. 

By comparison, LIME recognizes that the effect of word removal may not be linearly additive (i.e., the total prediction change from removing two words is not the sum of prediction changes from removing each word), and thus proposes to remove random subsets of words, computes the model prediction change, and attributes the change to individual words by solving a linear regression problem. 

\subsection{Evaluation Metrics}

For faithfulness evaluations of the self-explanations versus occlusion and LIME, we use five evaluation metrics: comprehensiveness \citep{deyoung2020eraser}, sufficiency \citep{deyoung2020eraser}, decision flip rate under the removal of most important token (\dfmit{}) \citep{chrysostomou2021improving}, minimum fraction of tokens needing to be removed in order to cause decision flip (\dffrac{}) \citep{serrano2019attention} and rank correlation with model prediction after word deletion (\rankdel) \citep{alvarez2018robustness}. Given a candidate attribution explanation to be evaluated, these metrics produce a score based on the model predictions (and their changes) on (1) the original input and (2) the input with some words removed according to the explanation. We follow the definition and implementation by \citet{zhou2022solvability}. As before, the model prediction is taken from the decimal point number in the model's textual output. During the evaluation, when multiple words have the same saliency value (a feature unique to these LLM-based explanations, elaborated in Section \ref{sec:qual-analysis}), we perform random tie-breaking among them to determine the word removal order. 

To evaluate the \topk{} explanations (which does not have any ranking among the top-$k$ words or among the remaining words), we adapt comprehensiveness, sufficiency and \dfmit{} to their \topk{} versions. Specifically, let $\vec x$ be the original input, $\vec x_{-k}$ be the input with the \topk{} words removed, and $\vec x_k$ be the input with only \topk{} words kept. Then we define
\begin{align}
    \text{comp@$k$}&=f(\vec x)-f(\vec x_{-w}), \\
    \text{suff@$k$}&=f(\vec x)-f(\vec x_{w}), \\
    \text{\dfmit{}@$k$}&=\mathbbm 1_{[f(\vec x)>0.5]\neq [f(\vec x_{-w})>0.5]}, 
\end{align}
where the indicator function in \dfmit{}@$k$ evaluates to 1 if removing all $k$ words together flips the label, and 0 otherwise. For non-\topk{} explanations, we can also evaluate them on these metrics by considering the $k$ features with highest explanation values (random tie-breaking if necessary) as the \topk{} features.

Besides these faithfulness evaluations (i.e., computing a metric score for an individual explanation), we also evaluate their (dis)agreement, motivated by \citet{disagreementProblem} and directly using their definitions. 

\begin{figure*}[b]
    \centering
    \includegraphics[width=\textwidth]{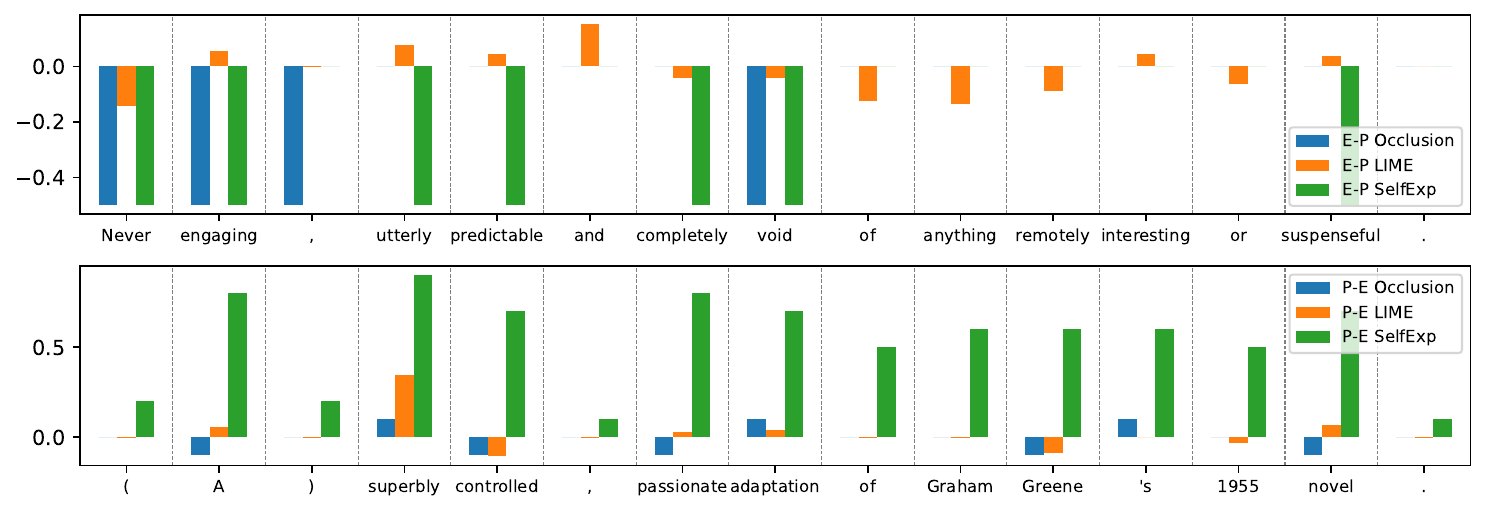}
    \caption{Visualization of one explanation each for E-P and P-E model. The \topk{} explanations}
    \label{fig:sample-explanation}
\end{figure*}

\subsection{A Caveat on Model Sameness}
Since the model is auto-regressive in nature, its output is affected by two factors: the model weight and the context (i.e., the prompt). Thus, given the same model (e.g., ChatGPT), prompting it in different ways (e.g., via different system messages) effectively leads to different models. Thus, in our experiment, we need to treat the E-P and P-E prompt as two different models, and the generated self-explanations are not comparable to each other. In addition, in order to make meaningful comparisons between self-explanation and occlusion/LIME, we need the latter to be generated with the same system prompt, too. 

Concretely, as an example, for the sentence ``\textit{I really like this movie}'', to obtain the occlusion value for the word ``like'' that is comparable to E-P self-explanation, we need to compute the model prediction difference on the input ``\textit{I really like this movie}'' vs. ``\textit{I really this movie}'', \textit{both using the prompt as shown in Table \ref{tab:ep-prompt}, even though we do not need the explanation value for the latter sentence.} The same reasoning applies on the evaluation side, where in order to retrieve the model prediction values, we need to use the original prompt and ignore the explanation output byproduct. 

This also means that the \topk{} E-P and P-E explanations are not those for the E-P and P-E models. However, there is no way around this issue: in order for the model to generate the \topk{} explanation, it has to be instructed in that way, which effectively results in a different model. Thus, in our study, we ignore the difference, and consider the \topk{} explanation effectively as an explanation generated by an external explainer, for the target E-P or P-E model.

\section{Experiments and Results}

\subsection{Experiment Setup and Explanation Visualization}
We conduct experiments on the Stanford Sentiment Treebank (SST) dataset \cite{socher-etal-2013-recursive}, which is a collection of short- to medium-length movie reviews, with their associated labels. While the labels are provided as a continuous value between 0.0 (totally negative) and 1.0 (totally positive), we binarize the label at 0.5. Due to the cost and rate limit of the OpenAI's API access which makes the full set of experiments on a single instance taking around half an hour, we select a random subset of 100 sentences from the test set for our investigations.

With this setup, Figure \ref{fig:sample-explanation} presents the explanations for two sentences. We see that the LLM-generated self-explanations are intuitively reasonable in highlighting words of strong intrinsic sentiment values such as ``superbly'' and ``void''. In addition, there are other features, such as that the explanation values are quite sparse and distributed on a few ``levels,'' which are discussed further in Section \ref{sec:qual-analysis}. 

\subsection{Model Accuracy}

\begin{table}[t]
    \centering
    \caption{Prediction accuracy for models evaluated in the experiments.}
    \label{tab:accuracy-summary}
    \begin{tabular}{lr|lr}\toprule
        \multicolumn{2}{l}{Prediction Only} & \multicolumn{2}{r}{92\%} \\\midrule
        E-P & 85\% & P-E & 88\% \\
        E-P top-$k$ & 80\% & P-E top-$k$ & 83\% \\\bottomrule
    \end{tabular}
\end{table}

Table \ref{tab:accuracy-summary} summarizes the classification accuracy of each model. First, we note that the classification accuracy of E-P is lower than that of P-E, at 85\% vs 88\%. This difference suggests that generating an explanation before making the prediction is not always helpful, unlike the benefits observed in chain-of-thought prompting \citep{chainOfThoughtWReasoning}. We believe that this difference is due to the nature of the tasks: the ``reasoning'' in sentiment analysis cannot be adequately represented by attribution scores assigned to each word, and hence forcing the model to make a prediction based on preceding explanation (i.e., the E-P setup) hurts the performance. 

Furthermore, both models have worse performance compared to the model that does not need to generate any explanation, which has an accuracy of 92\%, with the system prompt shown below. 

\vspace{0.08in}
\noindent
\fbox{%
    \parbox{0.96\columnwidth}{
    \textit{\noindent
You are a creative and intelligent movie review analyst, whose purpose is to aid in sentiment analysis of movie reviews. A review will be provided to you, and you must classify the review as either 1 (positive) or 0 (negative), as well as your confidence in the score you chose. The confidence should be a decimal number between 0 and 1, with 0 being the lowest confidence and 1 being the highest confidence. Output this in the Python tuple format (<int classification>, <float confidence>).
\newline\newline
The movie review will be surrounded by <review> tags.
\newline\newline
Example output:
(<int classification>, <float confidence>)}
    }
}
\vspace{0.08in}

This result is reminiscent of that by \citet{eSNLINLEs}, where asking the model to provide explanations hurts the performance, despite with different mechanisms: prompting in our work and supervised training in the other work.

Last, the prediction is even less accurate when it needs to be made based on a few important words (in the E-P \topk{} setting), or when it needs to be justified later with a few words (in the P-E \topk{} setting). All thse model accuracy results imply that feature attribution explanations may not be the best form of explanation for sentiment analysis, which forces the model into an uncomfortable accuracy-interpretability trade-off \citep{gunning2019darpa}. 

\subsection{Faithfulness Evaluation Results}

Table \ref{tab:result-table} presents the performance of \selfexp{} for E-P and P-E models, compared to occlusion and LIME generated for each model respectively. We make several remarks below. 
\begin{table}[t]
    \centering
    \caption{Comparison between the self-explanation generated by ChatGPT and traditional explanations computed on it.}
    \label{tab:result-table}
    \begin{tabular}{r|ccccc}\toprule& 
        Comp $\uparrow$ & Suff $\downarrow$ & \dfmit{} $\uparrow$ & \dffrac{} $\downarrow$ & \rankdel{} $\uparrow$\\\midrule
         & \multicolumn{5}{c}{E-P (Accuracy: 85\%)}\\
         Occlusion & 0.15 & 0.26 & \textbf{0.18} & 0.57 & \phantom{-}\textbf{0.00}\\
         LIME & 0.17 & \textbf{0.22} & 0.13 & \textbf{0.50} & -0.02 \\
         \selfexp{} & \textbf{0.19} & 0.25 & 0.16 & 0.55 & -0.03\\\midrule
         & \multicolumn{5}{c}{P-E (Accuracy: 88\%)}\\
         Occlusion & 0.20 & 0.23 & \textbf{0.14} & 0.64 & -0.02 \\
         LIME & \textbf{0.27} & \textbf{0.20} & 0.10 & \textbf{0.56} & \phantom{-}\textbf{0.02} \\
         \selfexp{} & \textbf{0.27} & 0.22 & 0.07 & \textbf{0.56} & -0.01
         \\\bottomrule
    \end{tabular}
\end{table}

First, we note the lack of a clear winner across the different metrics. Unexpectedly, we see that the occlusion saliency performs the best on \dfmit{}, because its definition (i.e., the importance of a word defined as the model prediction change under word removal) inherently makes its most important word being the most likely to flip the model prediction when removed -- an observation of solvability made by \citet{zhou2022solvability}. We should also expect it to achieve the first place on \rankdel{}, but all the numbers are very close to 0 and, as we will show in Section \ref{sec:qual-analysis}, the insensitivity of model prediction under single word removal makes this metric highly prone to the effect of stochasticity in tie-breaking. 

Among the remaining metrics, occlusion is the best on none, which is not surprising as it could not capture the non-linear effects of feature removals. Such non-linear effects are considered by LIME, which performs quite evenly with \selfexp{}. 

Nonetheless, running LIME is highly expensive: since it needs to solve a linear regression problem, the default number of perturbations by the Python \texttt{lime} package\footnote{\url{https://lime-ml.readthedocs.io/en/latest/}} is 5,000, meaning that we need to query the OpenAI API 5,000 times in order to generate the explanation for a single sentence. In practice, we adaptively scale this number to be 10 times the number of tokens in the sentence with an average of 183 perturbations per explanation, but this still is quite expensive, taking approximately 20 minutes per explanation.

Thus, given that \selfexp{} performs on par with LIME, we see that it could be low-cost substitute of LIME, as this self-explanation is generated along with the prediction.

As mentioned in the Section \ref{sec:method-prompting}, while the Python-like format guiding helps the model to generate syntactically valid explanation and prediction, we are concerned that the explanation quality may suffer. Thus, we also prompt the model to generate the top-$k$ most important words, and without attribution values attached to them. Table \ref{tab:result-table-topk} shows three metrics that are compatible with the top-$k$ explanation, and to compare it with other (fully formed) explanations, we extract the top-$k$ words from the latter.

\begin{figure*}[b]
    \centering
    \includegraphics[width=\textwidth, trim={3cm 1.5cm 3cm 1.5cm}, clip]{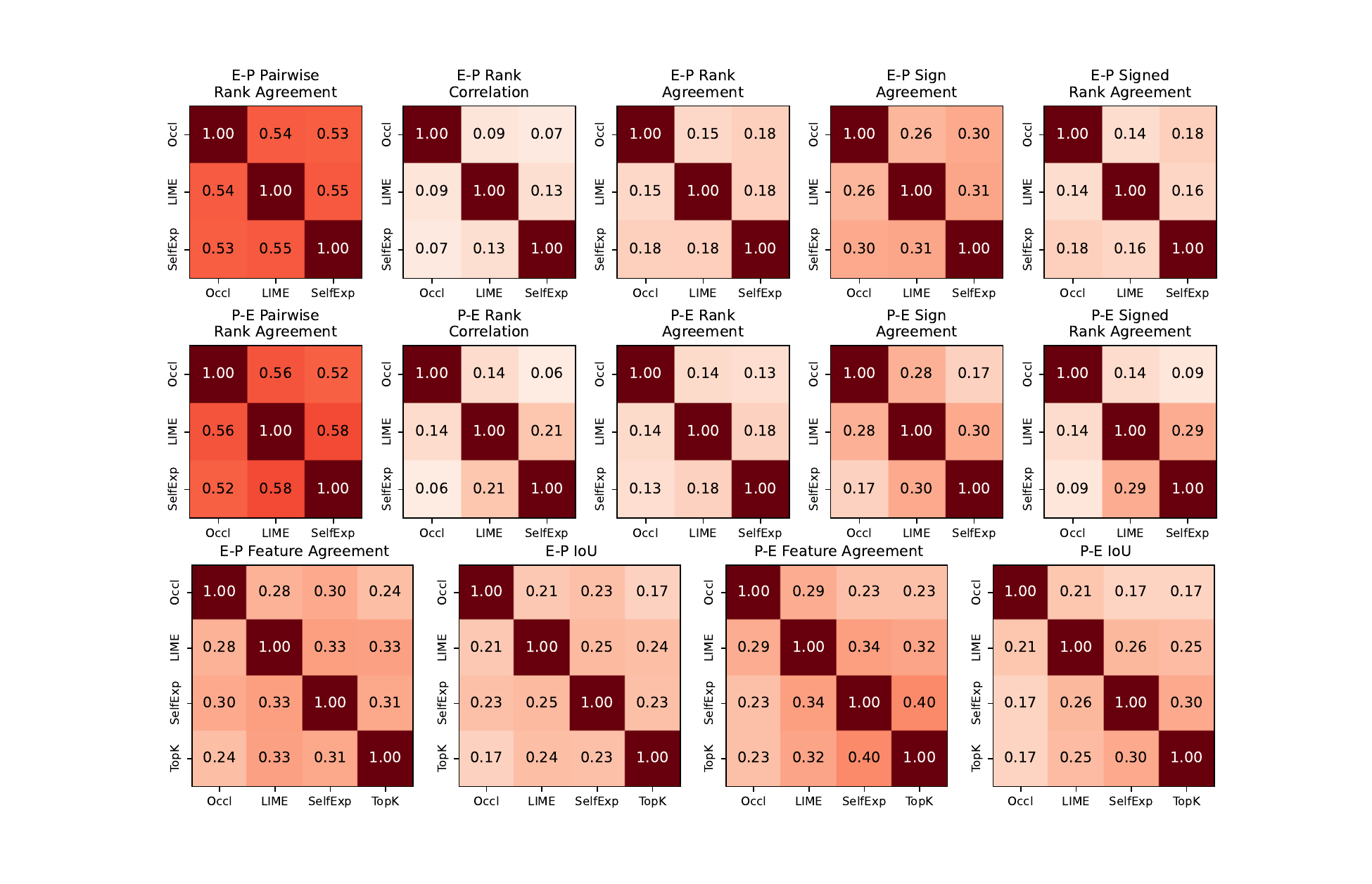}
    \caption{The agreement metric values among different explanations.}
    \label{fig:agreement}
\end{figure*}

\begin{table}[t]
    \centering
    \caption{Comparison of all methods with the \topk{} explanations on \topk{}-adapted metrics.}
    \label{tab:result-table-topk}
    \begin{tabular}{r|ccc}\toprule
         & Comp@$k$ $\uparrow$ & Suff@$k$ $\downarrow$ & \dfmit{}@$k$ $\uparrow$\\\midrule
         & \multicolumn{3}{c}{E-P (Accuracy: 80\%)}\\
         Occlusion & 0.16 & 0.34 & 0.29 \\
         LIME & \textbf{0.17} & 0.28 & \textbf{0.40} \\
         \selfexp{} & 0.11 & 0.32 & 0.32 \\
         \selfexpk{} & 0.10 & \textbf{0.27} & 0.34 \\\midrule
         & \multicolumn{3}{c}{P-E (Accuracy: 83\%)}\\
        Occlusion & 0.11 & 0.29 & 0.17\\
        LIME & \textbf{0.18} & 0.26 & \textbf{0.31}\\
        \selfexp{} & 0.11 & 0.30 & 0.29\\
        \selfexpk{} & 0.11 & \textbf{0.25} & \textbf{0.31}\\
         \bottomrule
    \end{tabular}
\end{table}

As we can see, the \topk{} explanation is not distinctively better -- while it achieves the best result on some metrics, it (or any other explanation) is by no means the clear winner. Thus, given that explanations in the wild are often in the \topk{} format (c.f., Table \ref{tab:top-k-in-the-wild}), the results suggest caution to us when interpreting these explanations. 

\subsection{Agreement Evaluation Results}

Figure \ref{fig:agreement} shows the agreement among various explanations, for both the E-P and P-E setups. Similar to the findings by \citep{disagreementProblem}, high disagreement among every pair of explanations is exhibited. However, the faithfulness evaluation does not identify any method as a top performer across the board, but instead all methods perform very closely to each other. This is in stark contrast with traditional interpretability methods generated for classifiers, where the performance difference among different methods is usually much larger (c.f., results by \citet{zhou2022solvability}). In other words, while \citet{disagreementProblem} found that explanations that perform very differently on faithfulness metrics (for classifier models) have high disagreement, we find that explanations that perform similarly on faithfulness metrics also have high disagreement (for the ChatGPT model).

There could be two possible reasons for our finding. The first one is that our evaluations do not have the discriminative power to differentiate the good explanations from the bad ones. Some evidence supporting this claim is presented in Section \ref{sec:prediction-value-analysis} below. The second reason is that there could be some explanations that achieve distinctively higher faithfulness metric values, which understandably also has high disagreement with all of the current ones. In other words, we are only obtaining mediocre explanations with occlusion, LIME and self-explanation, while missing out on the ``optimal one.'' We leave further investigation of this topic to future work.

\subsection{Qualitative Analyses}
\label{sec:qual-analysis}

During our experiments, we notice several differences between the explanation generation and evaluation for ChatGPT versus those for other models. These differences often stem from fundamental differences between ChatGPT and other models, and serve as important caveats when generating and understanding these explanations. 

\subsubsection{Level of Saliency Values}
In the \selfexp{} attribution values, it is very common for them to be on a few levels, whose values are often ``well-rounded'' decimal numbers such as 0.5 and 0.75, rather than arbitrary ones such as 0.6423 and 0.1982 (often produced by traditional techniques for traditional models). One example is shown in Table \ref{tab:ep-prompt} and Table \ref{tab:pe-prompt}, and additional instances are presented in Table \ref{tab:leveled-saliencies}. Quantitatively, on average each sentence has 18.2 words, but there are only 2.72 levels of unique explanations per sentence for the E-P model and 5.69 for the P-E model. 

\begin{table}[!b]
    \centering
    \caption{Additional examples of E-P and P-E model responses.}
    \label{tab:leveled-saliencies}
    \begin{tabular}{p{8cm}}\toprule
         \textbf{E-P Prompting}\\\midrule
        <review> \textcolor{gray}{\textit{Ford deserves to be remembered at Oscar time for crafting this wonderful portrait of a conflicted soldier .}} <review>\newline
        \textcolor{Cerulean}{\textbf{[(`Ford', 0.500), (`deserves', 0.500), (`to', 0.000), (`be', 0.000), (`remembered', 0.500), (`at', 0.000), (`Oscar', 0.000), (`time', 0.000), (`for', 0.000), (`crafting', 0.500), (`this', 0.000), (`wonderful', 1.000), (`portrait', 0.500), (`of', 0.000), (`a', 0.000), (`conflicted', -0.500), (`soldier', 0.000), (`.', 0.000)]
        \newline (1, 0.800)}}\\\midrule
        <review> \textcolor{gray}{\textit{It 's never a good sign when a film 's star spends the entirety of the film in a coma .}} <review>\newline
        \textcolor{Cerulean}{\textbf{[(`It', 0.5), (``'s'', 0.5), (`never', -0.5), (`a', 0.0), (`good', 0.8), (`sign', 0.6), (`when', 0.0), (`a', 0.0), (`film', 0.0), (``'s'', 0.5), (`star', 0.4), (`spends', 0.2), (`the', 0.0), (`entirety', 0.0), (`of', 0.0), (`the', 0.0), (`film', 0.0), (`in', 0.0), (`a', 0.0), (`coma', -0.7), (`.', 0.0)]
        \newline (0, 0.700)}}\\\midrule
         \textbf{P-E Prompting}\\\midrule
         <review> \textcolor{gray}{\textit{Ford deserves to be remembered at Oscar time for crafting this wonderful portrait of a conflicted soldier .}} <review>\newline
        \textcolor{Cerulean}{\textbf{(1, 0.9)
        \newline [(`Ford', 0.2), (`deserves', 0.8), (`to', 0.1), (`be', 0.1), (`remembered', 0.7), (`at', 0.1), (`Oscar', 0.5), (`time', 0.2), (`for', 0.1), (`crafting', 0.6), (`this', 0.1), (`wonderful', 0.9), (`portrait', 0.7), (`of', 0.1), (`a', 0.1), (`conflicted', 0.8), (`soldier', 0.7), (`.', 0.1)]}}\\\midrule
        <review> \textcolor{gray}{\textit{Even die-hard fans of Japanese animation ... will find this one a challenge .}} <review>\newline
        \textcolor{Cerulean}{\textbf{(0, 0.8)
        \newline [(`Even', 0.2), (`die-hard', 0.1), (`fans', 0.3), (`of', 0.0), (`Japanese', 0.5), (`animation', 0.6), (`...', 0.0), (`will', 0.0), (`find', 0.0), (`this', 0.0), (`one', 0.0), (`a', 0.0), (`challenge', 0.4), (`.', 0.0)]}}
        \\\bottomrule
    \end{tabular}
\end{table}

We believe that this phenomenon occurs due to the human-like behavior and reasoning capability of ChatGPT. In its attempt to mimic a human (which is the central goal of the reinforcement learning from human feedback (RLHF) alignment \citep{ouyang2022training}), it outputs explanation values that are plausibly produced by humans, who will not give very fine-grained saliency values (e.g., 0.6423 and 0.1982) as these values are often beyond the human's reasoning ability.

As recent papers have noted \citep{zheng2021irrationality, zhou2022exsum}, model explanations are often less understandable to humans than desirable in that humans may not be able to draw the correct high-level conclusions about the model's reasoning from individual (i.e., local) explanations. One potential culprit could be the traditionally very fine-grained and varied saliency values, even if they faithfully reflect the model's reasoning. Thus, the ChatGPT self-explanation may offer a different kind of faithfulness-understandability trade-off, where the explanations are intentionally vague but easier for human consumption. Testing this hypothesis requires a formal evaluation of understandability with human subjects, which we leave to future work. Orthogonally, comparing the explanations generated by aligned and unaligned models could identify the impact of alignment, though getting access to the latter model may prove difficult as companies are often reluctant to release these models due to safety concerns \citep{irving2018ai}.

We attempted to combat the issue with few-shot prompting, where we provide some explanation values according to our intuitions on some examples before asking the model to generate its own explanation and prediction. However, the results are less than satisfactory. Few-shot prompting does solve the issue of saliency values being too ``rounded:'' when provided with more fine-grained values (e.g. 0.8694 and -0.1285) as attribution scores, the generated explanations also tend to be less rounded. However, still not too many saliency levels are present. In addition, ChatGPT tends to ``memorize'' the provided attribution values for specific words. For example, when the word ``the'' has saliency value of 0.163 in the exemplar, the model is likely to assign the same value of 0.163 to other instances of ``the'' in the query sentence. A full example demonstrating both phenomena is presented in Table \ref{tab:few-shot-prompting}. As a result of the weaknesses of the few-shot prompting, we opted to base our experiments on the more stable, human-like responses from the zero-shot prompting setup. Overall, our findings suggest that the model is learning to mimic the human-provided explanation, which is a fundamentally flawed behavior: we want the model's explanation to reflect its own reasoning process, rather than reflecting the reasoning process of another decision maker (human or model). 

\begin{table*}[!htb]
    \centering
    \caption{An example few-shot learning conversation. All texts up to and including the last user message are human generated and provided as instructions and few-shot exemplars. Only the last assistant message in \textbf{boldface} is model-generated. \textcolor{ForestGreen}{Green} and \textcolor{Mulberry}{purple} colors show that the model still produces attribution values on few saliency levels, though they are no longer well-rounded. \textcolor{Cerulean}{Blue} color shows that the model memorizes the provided (word, attribution value) pair. }
    \label{tab:few-shot-prompting}
    \begin{tabular}{lp{16cm}}\toprule Role & Text \\\midrule 
    System & You are a movie review bot. Your job is to analyze movie reviews and classify them as either 1 (positive) or 0 (negative), as well as your confidence in the score you chose. Please also provide the degree of importance of each word in the sentence in your explanation in the format ("word", importance). For example, the expected response for the sentence "This movie was very good" is: "Classification: 1, 0.99 confidence. [("This", 0.085), ("movie", 0.149), ("was", -0.177), ("very", 0.691), ("good", 0.997)]. " It does not matter whether or not the sentence makes sense. Do your best given the sentence.
    \newline\newline
    It is imperative that you respond in the desired format regardless of the validity of the sentence.
    \\\midrule User & Reggio 's trippy , ambitious downer can also sometimes come across like nothing more than a glorified Nike ad .
    \\\midrule Assistant & Classification: 0, 0.82 confidence. [("Reggio", 0.254), ("'s", 0.192), ("trippy", -0.392), (",", -0.045), ("ambitious", 0.498), ("downer", -0.602), ("can", 0.195), ("also", 0.075), ("sometimes", 0.285), ("come", 0.043), ("across", 0.177), ("like", 0.101), ("nothing", -0.255), ("more", -0.101), ("than", 0.121), ("a", 0.004), ("glorified", 0.384), ("Nike", -0.369), ("ad", -0.739), (".", 0.007)]
    \\\midrule User & There is not a single movie that could have been better than this .
    \\\midrule Assistant & Classification: 1, 0.90 confidence. [("There", 0.004), ("is", 0.114), ("not", -0.787), ("a", 0.119), ("single", 0.239, ("movie", 0.395, ("that", 0.043), ("could", 0.294), ("have", 0.155), ("been", 0.020), ("better", 0.859), ("than", 0.122), ("this", 0.500), (".", 0.001)]
    \\\midrule User & It was a great movie overall , but the ending was a bit lackluster .
    \\\midrule Assistant & Classification: 1, 0.75 confidence. [("It", 0.174), ("was", -0.101), ("a", 0.122), ("great", 0.825), ("movie", 0.608), ("overall", 0.390), (",", -0.009), ("but", -0.134), \textcolor{Cerulean}{("the", 0.033)}, ("ending", -0.635), ("was", -0.145), ("a", 0.103), ("bit", -0.396), ("lackluster", -0.859), (".", -0.003)]
    \\\midrule User & The film provides some great insight into the neurotic mindset of all comics even those who have reached the absolute top of the game .
    \\\midrule  Assistant & \textbf{Classification: 1, 0.98 confidence. [\textcolor{Cerulean}{("The", 0.033)}, ("film", 0.607), ("provides", 0.346), ("some", 0.091), ("great", 0.825), \textcolor{ForestGreen}{("insight", 0.537)}, ("into", 0.091), \textcolor{Cerulean}{("the", 0.033)}, \textcolor{ForestGreen}{("neurotic", -0.498)}, \textcolor{ForestGreen}{("mindset", -0.498)}, \textcolor{Mulberry}{("of", 0.033)}, \textcolor{Mulberry}{("all", 0.033)}, \textcolor{ForestGreen}{("comics", 0.537)}, \textcolor{Mulberry}{("even", 0.033)}, \textcolor{Mulberry}{("those", 0.033)}, \textcolor{Mulberry}{("who", 0.033)}, \textcolor{Mulberry}{("have", 0.033)}, \textcolor{Mulberry}{("reached", 0.033)}, \textcolor{Mulberry}{("the", 0.033)}, \textcolor{Mulberry}{("absolute", 0.033)}, \textcolor{Mulberry}{("top", 0.033)}, \textcolor{Mulberry}{("of", 0.033)}, \textcolor{Mulberry}{("the", 0.033)}, \textcolor{Mulberry}{("game", 0.033)}, \textbf{(".", 0.000)}]} \\\bottomrule
    \end{tabular}
\end{table*}

\subsubsection{Prediction Value Roundedness and Insensitivity}
\label{sec:prediction-value-analysis}
A key component in both the definition and evaluation of model explanations is the model's prediction on partial sentences with certain words removed. Recall that since we do not have access to token probability output from ChatGPT, we define the model prediction as the textual output of the confidence score as a floating point number. Since this textual output is also shaped by alignment, we have similar findings as before. Specifically, we notice that the model's prediction values are often ``well-rounded'' and being one of several distinct values, as shown in Table \ref{tab:leveled-prediction}.

\begin{table}[!b]
    \centering
    \caption{A random selection of input sentences and their predictions by the E-P model (similar results for other models).}
    \label{tab:leveled-prediction}
    \begin{tabular}{p{7cm}|c}\toprule
        Sentence & Pred \\\midrule
        Guided more by intellect than heart , his story flattens instead of sharpens . & 0.70\\\midrule
        These are names to remember , in order to avoid them in the future . & 0.50\\\midrule
        A hideous , confusing spectacle , one that may well put the nail in the coffin of any future Rice adaptations . & 0.00\\\midrule
        Eastwood is an icon of moviemaking , one of the best actors , directors and producers around , responsible for some excellent work . & 0.90\\\midrule
        While Super Troopers is above Academy standards , its quintet of writers could still use some more schooling . & 0.80\\\bottomrule
        
    \end{tabular}
\end{table}

\begin{table}[!b]
    \centering
    \caption{E-P model prediction on a sequence of word removals (similar results for the P-E model). The word that will be removed in the next step is shown in \textcolor{Cerulean}{\textbf{blue boldface}}. }
    \label{tab:successive-removal-prediction}
    \begin{tabular}{p{7cm}|c}\toprule
        Sentence & Pred \\\midrule
        One of the greatest family-oriented , fantasy-adventure movies \textcolor{Cerulean}{\textbf{ever}} . & 0.90\\\midrule
        One of the \textcolor{Cerulean}{\textbf{greatest}} family-oriented , fantasy-adventure movies . & 0.90\\\midrule
        One of the \textcolor{Cerulean}{\textbf{family-oriented}} , fantasy-adventure movies . & 0.80\\\midrule
        One of the, \textcolor{Cerulean}{\textbf{fantasy-adventure}} movies . & 0.80\\\midrule
        \textcolor{Cerulean}{\textbf{One}} of the , movies . & 0.80\\\midrule
        of the , movies \textcolor{Cerulean}{\textbf{.}} & 0.50\\\midrule
        of the , \textcolor{Cerulean}{\textbf{movies}} & 0.50\\\midrule
        \textcolor{Cerulean}{\textbf{of}} the , & 0.50\\\midrule
        \textcolor{Cerulean}{\textbf{the}} , & 0.50\\\midrule
        , & 0.50\\\bottomrule
    \end{tabular}
\end{table}

Furthermore, the removal of a few words often does not change the model's prediction, as Table \ref{tab:successive-removal-prediction} shows. This phenomenon has two causes: first, the model could likely infer the missing words and treat them as minor typographical errors; second, the model is aligned to human thinking, which habitually ignores these errors when making predictions. 

This model behavior has two implications. First, the occlusion saliency for a word is defined as the model prediction difference on inputs with and without the word, which results in the saliency value being exactly 0 for 82.6\% of words for the E-P model and 54.5\% for the P-E model. 

More importantly, all evaluations are affected to different extents by this behavior. The most severely affected ones are \dfmit{} and \rankdel{}, which are computed from the model prediction on the original sentence and that of one-word-removed sentence only, similar to occlusion saliency. As a result, all explanations have very low scores on these two metrics, compared to the same evaluation settings on a finetuned RoBERTa model \citep{liu2019roberta} which has much more gradual model prediction changes under word removal ($\sim 25\%$ for \dfmit{} and $\sim 0.5\%$ for \rankdel{} on average as reported by \citep{zhou2022solvability}). Notably, we did not observe a perfect 1.0 score for occlusion on \rankdel{}, because many saliency values and word removal prediction changes are the same (at which point random tie-breaking is applied). Even with more words being removed, the model's prediction still stays the same for a long time, which decreases the comprehensiveness score, as the initial few word removals could not significantly decrease the model's confidence, and increases the sufficiency score, as the initial few word insertions could not significantly increase the model's confidence.

\section{Conclusion and Future Work}

In this paper, we conducted the first rigorous assessment in LLM's capability to self-generate feature attribution explanations -- i.e., explaining its own prediction by stating the importance of every word in the input. Using ChatGPT and the SST dataset as our testbed, we construct prompts that reliably generate two types of LLM-generated self-explanations: full feature attribution explanation that assigns every word an importance score and a \topk{} explanation that only highlights the few most important words. The explanations are generated in two settings: one where they are generated first and then the model predictions (i.e., sentiment assessment) is made based on them, and the other where the prediction is generated first and the explanations are generated to justify the predictions. We compare these explanations to traditional explanation techniques -- occlusion saliency and LIME -- on both faithfulness and agreement metrics. 

We have two major findings from our investigation. First, none of the explanations, self-generated or not, hold a distinct advantage over the rest according to the faithfulness evaluations, yet they are highly different according to the agreement evaluations. As a result, there could potentially be some explanation which is much better than the current ones, and novel techniques may be needed to reveal them. 

Second, and possibly of independent interest in the study of LLMs, both the model prediction values and word attribution values are highly well-rounded, taking values such as 0.25, 0.67, 0.75, etc. This finding may also explain the similarity in faithfulness metric values in across different explanations: due to the lack of fine-grained variations in the explanation and prediction values, the evaluation metrics may not have the sufficient power to distinguish the good explanations from the bad ones. 

These findings also inform directions for future work. First, given that there is no prior work on studying LLM-generated feature attribution explanations, it is likely that our solution is not optimal, and better ways to elicit self-explanations could be developed. Second, we may need to rethink the evaluation practice, which has shown to not perform very well. Indeed, if we consider ChatGPT's reasoning ability as being more similar to humans than to earlier machine learning models, the result is not surprising: a human will not be able to give a precise number for the effect of removing one word in a sentence of twenty words (as used by \dfmit{} and \rankdel{} metrics), or give a list of the cumulative influence of a word deletion or insertion sequence (as used by comprehensiveness, sufficiency and \dffrac{} metrics). Thus, alternative evaluation strategies, possibly involving carefully designed human subject studies \citep[e.g.][]{bansal2021does} are needed to definitively assess the strengths and weaknesses of these explanations. 

An orthogonal direction would be to evaluate other LLMs, such as GPT-4, Bard and Claude. A comparative study could reveal important insights in how these models understand themselves (i.e., introspection), as explanations are essentially properties about predictions (e.g., occlusion saliency is the prediction change under single word removal). In addition, we can conduct the analogous studies on other types of explanations, such as counterfactual explanations \citep{wachter2017counterfactual} and concept-based explanations \citep{kim2018interpretability, li2018deep}. Finally, now that we have demonstrated that these explanations are feasible options to consider, we need to make sure that they are indeed beneficial and not used for harmful purposes. For example, how easily can they be manipulated \citep{slack2020fooling}, and could they hide fairness issues of the model \citep{dai2022fairness}? Identifying and solving any observed issues is a necessary step before the deployment of these explanations.

\bibliographystyle{plainnat}
\bibliography{big_bib, references}

\end{document}

%% file: tables/ep.tex
    \centering
    \caption{The prompt and an example query of the E-P setup. Black text in normal font is provided literally. \textcolor{gray}{\textsl{Gray text in italics}} is changed for every query. \textcolor{Cerulean}{\textbf{Blue text in boldface}} is the model's output.}
    \label{tab:ep-prompt}
    \begin{tabular}{lp{16cm}}\toprule
    Role & Text \\\midrule
    System & You are a creative and intelligent movie review analyst, whose purpose is to aid in sentiment analysis of movie reviews. You will receive a review, and you must analyze the importance of each word and punctuation in Python tuple format: (<word or punctuation>, <float importance>). Each word or punctuation is separated by a space. The importance should be a decimal number to three decimal places ranging from -1 to 1, with -1 implying a negative sentiment and 1 implying a positive sentiment. Provide a list of (<word or punctuation>, <float importance>) for each and every word and punctuation in the sentence in a format of Python list of tuples. Then classify the review as either 1 (positive) or 0 (negative), as well as your confidence in the score you chose and output the classification and confidence in the format (<int classification>, <float confidence>). The confidence should be a decimal number between 0 and 1, with 0 being the lowest confidence and 1 being the highest confidence.
    \newline\newline
    It does not matter whether or not the sentence makes sense. Do your best given the sentence.
    \newline\newline
    The movie review will be encapsulated within <review> tags. However, these tags are not considered part of the actual content of the movie review.
    \newline\newline
    Example output:
    \newline
    [(<word or punctuation>, <float importance>), (<word or punctuation>, <float importance>), ... ]
    \newline
    (<int classification>, <float confidence>)\\\midrule
    User & <review> \textcolor{gray}{\textit{Offers that rare combination of entertainment and education .}} <review>\\\midrule
    Assistant & \textcolor{Cerulean}{\textbf{[(`Offers', 0.500), (`that', 0.000), (`rare', 0.500), (`combination', 0.000), (`of', 0.000), (`entertainment', 0.750), (`and', 0.000), (`education', 0.750), (`.', 0.000)]
    \newline
    (1, 1.000)}}\\\midrule
    \end{tabular}
\end{table*}

%% file: tables/pe.tex
    \centering
    \caption{The prompt and an example query of the P-E setup. Black text in normal font is provided literally. \textcolor{gray}{\textit{Gray text in italics}} is changed for every query. \textcolor{Cerulean}{\textbf{Blue text in boldface}} is the model's output.}
    \label{tab:pe-prompt}
    \begin{tabular}{lp{16cm}}\toprule
    Role & Text \\\midrule
    System & You are a creative and intelligent movie review analyst, whose purpose is to aid in sentiment analysis of movie reviews. A review will be provided to you, and you must classify the review as either 1 (positive) or 0 (negative), as well as your confidence in the score you chose. The confidence should be a decimal number between 0 and 1, with 0 being the lowest confidence and 1 being the highest confidence. Output this in the Python tuple format (<int classification>, <float confidence>).
    \newline\newline
    Then, analyze how important every single word and punctuation token in the review was to your classification. The importance should be a decimal number to three decimal places ranging from -1 to 1, with -1 implying a negative sentiment and 1 implying a positive sentiment. Provide a list of (<word or punctuation>, <float importance>) for each and every word and punctuation token in the sentence in a format of Python list of tuples. Each word or punctuation is separated by a space.
    \newline\newline
    It does not matter whether or not the sentence makes sense. Do your best given the sentence.
    \newline\newline
    The movie review will be encapsulated within <review> tags. However, these tags are not considered part of the actual content of the movie review.
    \newline\newline
    Example output:
    \newline
    (<int classification>, <float confidence>)
    \newline
    [(<word or punctuation>, <float importance>), (<word or punctuation>, <float importance>), ... ]\\\midrule
    User & <review> \textcolor{gray}{\textit{A film that takes you inside the rhythms of its subject : You experience it as you watch .}} <review>\\\midrule
    Assistant & \textcolor{Cerulean}{\textbf{(1, 0.8)
    \newline
    [(`A', 0.2), (`film', 0.5), (`that', 0.1), (`takes', 0.3), (`you', 0.4), (`inside', 0.6), (`the', 0.1), (`rhythms', 0.7), (`of', 0.1), (`its', 0.1), (`subject', 0.5), (`:', 0.1), (`You', 0.4), (`experience', 0.6), (`it', 0.3), (`as', 0.2), (`watch', 0.4), (`.', 0.1)]}}\\\midrule
    \end{tabular}
\end{table*}

%% file: tables/wild.tex
\centering
    \caption{A ChatGPT response with human ``in the wild,'' where ChatGPT automatically gives the explanation in the form of a few highlighted phrases.}
    \label{tab:top-k-in-the-wild}
    \begin{tabular}{lp{16cm}}\toprule
    Role & Text \\\midrule
    User & Can you tell me whether the movie review is positive or negative? \newline
The gorgeously elaborate continuation of "The Lord of the Rings" trilogy is so huge that a column of words can not adequately describe co-writer/director Peter Jackson 's expanded vision of J.R.R. Tolkien 's Middle-earth .\\\midrule
System & Based on the given text, it appears to be a positive movie review. The phrases "gorgeously elaborate," "expanded vision," and "Middle-earth" suggest admiration and appreciation for the film.\\\bottomrule
    \end{tabular}
\end{table*}

%% file: tables/epk.tex
    \centering
    \caption{The prompt and an example query of the E-P top-$k$ setup. Black text in normal font is provided literally. \textcolor{gray}{\textit{Gray text in italics}} is changed for every query. \textcolor{Cerulean}{\textbf{Blue text in boldface}} is the model's output.}
    \label{tab:ep-prompt-topk}
    \begin{tabular}{lp{16cm}}\toprule
    Role & Text \\\midrule
    System & 
    As a movie review analyst, your role is to analyze the sentiment of movie reviews and provide insights on the importance of each word and punctuation in determining the overall positivity level. Your task is to identify the top \textcolor{gray}{\textit{3}} most significant words, ranked from the most positive sentiment to the least positive sentiment. Additionally, you need to determine whether the movie review is positive or negative along with your confidence in your prediction. A positive review is represented by the number 1, while a negative review will be represented by the number 0. The confidence should be a decimal score between 0 and 1, with 0 being the lowest confidence and 1 being the highest confidence. Please note that the coherence of the sentence is not relevant; your focus should be on analyzing the sentiment.
    \newline\newline
    The movie review will be enclosed within <review> tags, but these tags should not be included in the evaluation of the review's content.
    \newline\newline
    Only output the list of \textcolor{gray}{\textit{3}} words in the form of a comma separated list, with the prediction(as a number) and confidence following after, nothing more.\\\midrule
    User & <review> \textcolor{gray}{\textit{A film that takes you inside the rhythms of its subject : You experience it as you watch .}} <review>\\\midrule
    Assistant & \textcolor{Cerulean}{\textbf{rhythms, experience, watch, 1, 0.9}}\\\bottomrule
    \end{tabular}
\end{table*}

%% file: tables/pek.tex
    \centering
    \caption{The prompt and an example query of the P-E top-$k$ setup. Black text in normal font is provided literally. \textcolor{gray}{\textit{Gray text in italics}} is changed for every query. \textcolor{Cerulean}{\textbf{Blue text in boldface}} is the model's output.}
    \label{tab:pe-prompt-topk}
    \begin{tabular}{lp{16cm}}\toprule
    Role & Text \\\midrule
    System & 
    As a movie review analyst, your role is to analyze the sentiment of movie reviews and provide insights on the importance of each word and punctuation in determining the overall positivity level. Your task is to determine whether the movie review is positive or negative along with your confidence in your prediction. A positive review is represented by the number 1, while a negative review will be represented by the number 0. The confidence should be a decimal score between 0 and 1, with 0 being the lowest confidence and 1 being the highest confidence. In addition, you need to identify the top \textcolor{gray}{\textit{3}} most significant words, ranked from the most positive sentiment to the least positive sentiment. Please note that the coherence of the sentence is not relevant; your focus should be on analyzing the sentiment.
    \newline\newline
    The movie review will be enclosed within <review> tags, but these tags should not be included in the evaluation of the review's content.
    \newline\newline
    Only output the prediction(as a number) and confidence, with the list of \textcolor{gray}{\textit{3}} words in the form of a comma separated list following after, nothing more.\\\midrule
    User & <review> \textcolor{gray}{\textit{A film that takes you inside the rhythms of its subject : You experience it as you watch .}} <review>\\\midrule
    Assistant & \textcolor{Cerulean}{\textbf{1, 0.8, rhythms, experience, watch}}\\\midrule
    \end{tabular}
\end{table*}